\documentclass[runningheads]{llncs}
\usepackage{marvosym}

\usepackage{amsmath}
\usepackage{xcolor}
\usepackage{graphicx}
\usepackage{float}
\usepackage{bookmark}
\usepackage{booktabs}
\usepackage[ruled,linesnumbered]{algorithm2e}
\usepackage{algorithmic}
\usepackage{bm}
\usepackage{subfigure}
\usepackage{multirow}
\usepackage{hyperref}
\hypersetup{
    colorlinks=true,
    linkcolor=blue,
    urlcolor=blue,
    pdftitle={Residual Network and Embedding Usage: New Tricks of Node Classification with Graph Convolutional Networks},
    bookmarksopen=true,
}

\newcommand{\std}[1]{\textcolor{gray}{\scriptsize{$\pm$#1}}}

\begin{document}
\title{Residual Network and Embedding Usage:\\ New Tricks of Node Classification with Graph Convolutional Networks}
\titlerunning{Residual Network and Embedding Usage}
\author{Huixuan Chi\inst{1} \and Yuying Wang\inst{1} \and
Qinfen Hao\inst{1}\textsuperscript{(\Letter)} \and Hong Xia\inst{2}}

% \author{First Author \and
% Second Author\textsuperscript{(\Letter)} \and Third Author \and Fourth Author}
%
\authorrunning{Chi et al.}


\institute{
Institute of Computing Technology, Chinese Academy of Sciences
\email{chihuixuan@163.com},  \email{\{wangyuying,haoqinfen\}@ict.ac.cn}\\ 
\and School of Control and Computer Engineering, North China Electric Power University 
% \email{summerday@ncepu.edu.cn}
}

\maketitle              % typeset the header of the contribution

\begin{abstract}
\bookmark[dest=\HyperLocalCurrentHref,level=1]{Abstract}

Graph Convolutional Networks (GCNs) and subsequent
variants have been proposed to solve tasks on graphs, especially node
classification tasks. In the literature, however, most tricks or
techniques are either briefly mentioned as implementation details or
only visible in source code. In this paper, we first summarize some
existing effective tricks used in GCNs mini-batch training. Based on
this, two novel tricks named \textsl{GCN\_res Framework} and \textsl{Embedding
Usage} are proposed by leveraging residual network and pre-trained embedding to improve baseline's test accuracy in
different datasets. Experiments on Open Graph Benchmark (OGB) show that,
by combining these techniques, the test accuracy of various GCNs increases by 1.21\%$\sim$2.84\%. 
We open source our implementation at
 \url{https://github.com/ytchx1999/PyG-OGB-Tricks}.

\end{abstract}
%
%
%
%%%Introduction
\section{Introduction}\label{introduction}
\bookmark[dest=\HyperLocalCurrentHref,level=1]{1 Introduction}

More recently, Graph Convolutional Networks (GCNs) and their variants have successfully extended
convolution and pooling operation to graphs and achieved good
performance in many fields such as computer vision, recommendation
systems and natural language processing~\cite{wu2020comprehensive}. Currently, spatial-based GCNs, such as GCN \cite{kipf2016semi}, GraphSAGE \cite{hamilton2017inductive} and
GraphSAINT \cite{zeng2019graphsaint}, have gradually replaced spectral-based GCNs in practice due to their
efficiency and flexibility. In order to solve the scalability problem of GCNs, researchers have proposed two kinds of mini-batch training algorithms:
sampling-based \cite{hamilton2017inductive,zeng2019graphsaint,chen2018fastgcn} and clustering-based \cite{chiang2019cluster}, to extend GCNs to large-scale graphs. Tasks in graph
representation learning can be divided into three categories: node
classification, link prediction, and graph classification. The node
classification task has become one of the most popular benchmarks
in GCNs due to its intuitiveness and simplicity.

Much of the work has aimed at the practice of GCNs, e.g., Open Graph Benchmark (OGB) \cite{hu2020open}, which has greatly promoted the development 
of GCNs. Before the release of OGB datasets and its leaderboard, GCNs
have not had a unified and universally-followed experimental protocol.
Different studies have used different dataset splitters and
evaluators, which have a negative impact on the fairness of different experiments \cite{shchur2018pitfalls,errica2019fair}. Moreover, small graph datasets used
in the early days, such as Cora, Citeseer and Pubmed, are far away from the real-world graphs, making it
difficult to transfer some tricks to
large-scale and real-world graphs. The above factors have led to the  tricks of GCNs not
receiving enough attention in the early research. Some tricks are
either simply mentioned in the literature or only visible in the source code.
Fortunately, since the release of OGB datasets and its leaderboard, the importance
of tricks has gradually emerged under relatively fair evaluation
standards and real-world graphs. Besides, the gains brought by tricks have sometimes exceeded
the gains brought by model architecture improvement. However, no one has summarized the tricks of GCNs, which are also not complete so
far. The inner relationships between these tricks are
not clear, which bring difficulties to the application and further
development.

\paragraph{\textbf{Present work.}} Firstly, we review the mini-batch training process and the existing effective tricks of GCNs which can often make training faster and
better for node classification tasks. Based on this, we propose two noval tricks for node
classification tasks: \textbf{GCN\_res Framework} and \textbf{Embedding Usage}. By
applying (i) adaptive residual connections and initial residual connections, and (ii)
softmax layer-aggregation, \textsl{GCN\_res Framework} can relieve
the over-smoothing problem. For example, on \texttt{ogbn-arxiv}, the test accuracy of our model is 
improved by 1.39\% compared to GCN. We also
compare our model with other
baselines such as DeeperGCN \cite{li2020deepergcn}, GCNII \cite{chen2020simple} to show the huge gains.
Furthermore, the effectiveness of \textsl{Embedding Usage} is
verified through experiments on \texttt{ogbn-mag}, \texttt{ogbn-products} and \texttt{ogbn-proteins}. For example, on \texttt{ogbn-mag}, it increases GraphSAINT's test accuracy by 2.15\%. Finally, we
also find that different combinations of tricks can
often improve various GCNs. By combining different strategies, GraphSAGE's test accuracy 
increases from 78.70\% to 81.54\% on \texttt{ogbn-products}, achieving the best performance of the
model based on SAGEConv and its variants.

The main contributions of this research are as follows:
\begin{itemize}
\item[1)]
  We summarize the mini-batch training process and existing effective tricks of GCNs for node classification tasks.
\item[2)]
  We propose two novel tricks of GCNs for node classification tasks:  \textsl{GCN\_res Framework} and \textsl{Embedding Usage}, which can improve various GCNs significantly on OGB datasets\footnote{\url{https://ogb.stanford.edu/docs/nodeprop/}}. Besides, our work is also open sourced in OGB leaderboard\footnote{\url{https://ogb.stanford.edu/docs/leader_nodeprop/}}. 
\item[3)]
  Related work about tricks of GCNs is analyzed and discussed, and a new
  direction is proposed for the future work.
\end{itemize}

In Section \ref{preliminaries}, we will introduce the training process of GCNs and the
existing tricks. In Section \ref{methods}, we will present and discuss two novel tricks: \textsl{GCN\_res
Framework} and \textsl{Embedding Usage}. In Section \ref{experiments}, we will conduct our experiments and ablation studies on OGB datasets. In Section \ref{conclusion}, we will summarize this research.

%%% preliminary
\section{Preliminaries}\label{preliminaries}
\bookmark[dest=\HyperLocalCurrentHref,level=1]{2 Preliminaries}

\subsection{The Training Procedures of
GCNs}
\bookmark[dest=\HyperLocalCurrentHref,level=2]{2.1 The Training Procedures of
GCNs}

% In the following sections, we will use GCNs to represent the
% spatial-based \textsl{Graph Convolutional Networks}, and use GCN \cite{kipf2016semi} to represent the first spatial-based
% GCNs. 
The template of training GCNs with mini-batch
stochastic gradient descent is shown in Algorithm \ref{algo:minibatch}. Before initializing the model, we need to sample the whole graph to generate subgraphs required for mini-batch training first. In each iteration, we randomly select a subgraph to compute the gradients and then update the network parameters. It stops after N passes through the dataset. Finally, we can get the final node representation $\bm{Z}$ through a post-processing step 
of the base prediction. All functions
and hyper-parameters in Algorithm \ref{algo:minibatch} can be implemented
in many different ways.

\begin{algorithm}[h]
\caption{Train GCNs with mini-batch gradient descent.}
\label{algo:minibatch}
\begin{algorithmic}[1]
\REQUIRE{Graph $\mathcal{G}$; input feature matrix $\bm{X}$ }
\ENSURE{Final node representation $\bm{Z}$}
\STATE train\_loader $\leftarrow$ Sampling($\mathcal{G}$, $\bm{X}$);
\STATE initialize(model);
\FOR{epoch = 1, . . . ,N}
    \FOR{\emph{Sub\_Graph}\ \textbf{in}\ train\_loader}
        \STATE $out$ $\leftarrow$ forward(model, $\bm{X}$, \emph{Sub\_Graph});
        \STATE $loss$ $\leftarrow$ criterion($out$, y);
        \STATE $grad$ $\leftarrow$ backward($loss$);
        \STATE update(model, $grad$);
    \ENDFOR
\ENDFOR
\STATE $\bm{Z}$ $\leftarrow$ Post-processing($\mathcal{G}$, out);
\end{algorithmic}
\end{algorithm}

\subsection{Existing Efficient
Tricks}
\bookmark[dest=\HyperLocalCurrentHref,level=2]{2.2 Existing Efficient
Tricks}

\subsubsection{A. Mini-batch Training and
Sampling.} 
% ~\\
In large-scale graphs, it is quite difficult to train the whole graph
at a time in full-batch algorithms. Therefore, some researchers have
proposed sampling-based mini-batch algorithms to control the scale of graphs in the training process. The first
row and the fourth row in Algorithm \ref{algo:minibatch} are the steps for subgraph
sampling and subgraph loading.

There are some commonly used samplers such as random sampler, neighbor
sampler \cite{hamilton2017inductive} and GraphSAINT sampler \cite{zeng2019graphsaint}. In each batch, we only
need to train the subgraphs generated by sampling which not only saves the 
GPU memory, but also facilitates parallel computing across
multiple GPUs. Sampling can even be regarded as a kind of regularization method that randomly drops edges for nodes, so that the accuracy of GCNs can be slightly improved.

% Batch normalization technique \cite{ioffe2015batch} makes normalization a part of the model architecture and performing the normalization for each training mini-batch.
% This architecture design allows us to adopt much higher learning rates and be less careful about initialization.
% However, the effect of batch normalization is dependent on the mini-batch size and it is not obvious how to apply it to recurrent neural networks.
% To bridge this gap, layer normalization technique \cite{ba2016layer} is proposed to compute the normalization statistics separately at each time step, and it is very effective at stabilizing the hidden state dynamics in recurrent networks.

\subsubsection{B. Normalization and
Dropout.}
% ~\\
Normalization and dropout \cite{srivastava2014dropout} are widely used in CV, NLP and other fields.
Although they may be essential for training, they are rarely mentioned
in papers of GCNs. 

% \paragraph{\textbf{BatchNorm/LayerNorm or Not.}} 
How to normalize node features  is a very important issue while training. Many models will employ  BatchNorm (BN) \cite{ioffe2015batch} or LayerNorm (LN) \cite{ba2016layer} during training. However, normalization will also lead to a loss of node degree information and obscure various graph structure
features. Generally, Hamilton \textsl{et al.} \cite{hamilton2020graph} point out that normalization is most helpful in tasks where node feature information is much
more useful than structure information, or where there is a very wide range of node degrees. Both BN and LN can be written as
\begin{equation}
\bm{x}^{\prime}_i = \frac{\bm{x} -
        \textrm{E}[\bm{x}]}{\sqrt{\textrm{Var}[\bm{x}] + \epsilon}}
        \odot \gamma + \theta
\end{equation}
where $\textrm{E}[\bm{x}]$ and $\textrm{Var}[\bm{x}]$ stand for the mean and standard deviation respectively.

The BN/LN is very effective in GCNs, which can generally
stabilize the value, especially for deep GCNs. If there is
no BN/LN, then a numeric overflow may occur during training. Therefore, more and more GCNs
begin to use normalization. Recently, several new normalization methods, such as PairNorm \cite{zhao2019pairnorm} and NodeNorm \cite{zhou2020understanding}, are proposed to alleviate the over-smoothing problem in small graphs. Besides, normalization can also be used with dropout to prevent the overfitting problem.

% \paragraph{\textbf{Dropout or not.}} 
% Dropout can be used with normalization, to prevent overfitting.

\subsubsection{C. Adversarial Training
(FLAG).}
% ~\\
Standard adversarial training was first proposed to solve the min-max problem. Recently, many researchers are trying to
adopt adversarial training in GCNs for security purposes \cite{bojchevski2019adversarial,zhang2020gnnguard}. However, it remains unclear whether adversarial training can improve the accuracy of GCNs.

The FLAG \cite{kong2020flag} is a large-scale adversarial augmentation on graphs, which can tackle the overfitting problem by  adding gradient-based adversarial
perturbations to the input node features. Kong \textsl{et al.} \cite{kong2020flag} also point out that combining FLAG with normalization and
dropout can further improve the
performance of GCNs. 

\subsubsection{D. Label Propagation and
Usage.}
% ~\\
% \paragraph{\textbf{LP and C\&S.}} 
Some recent research attempts to connect label propagation with GCNs in order to utilize label information. Inspired by label propagation (LP) \cite{zhu2005semi}, 
Huang \textsl{et al.} \cite{huang2020combining} have proposed Correct and Smooth (C\&S) method, which corrects and smooths the base prediction through two
LPs. The C\&S method is often applied during the post-processing step in Algorithm \ref{algo:minibatch}.

% \paragraph{\textbf{Label Usage or not.}} 
Since the labels can
provide more information, Wang \textsl{et al.} \cite{wang2021bag} have proposed \textsl{Label Usage}, which uses a masking
technology to merge node features and labels as input. The motivation is similar to \textsl{Embedding Usage} that we are proposed
later, but in comparison, \textsl{Embedding Usage} is more
powerful.

%%% methods
% \vspace{-0.3cm}

% \vspace{-0.3cm}

\section{Proposed Methods}\label{methods}
\bookmark[dest=\HyperLocalCurrentHref,level=1]{3 Proposed Methods}
% In this section, we introduce our novel techniques working on the label and feature usages respectively.

\subsection{GCN\_res Framework and Residual
Network}
\bookmark[dest=\HyperLocalCurrentHref,level=2]{3.1 GCN\_res Framework and Residual
Network}

% \begin{figure*}[htbp]
%     \centering
%     \includegraphics[width=\textwidth]{GCN_res_fig.pdf}
%     \caption{Overview of \textsl{GCN\_res Framework} with a 4-layer toy example. The GCNs-Block consists of four parts: \texttt{GCNsConv} layer, \texttt{Norm} layer, activation function, and \texttt{Dropout} unit. Data stream of residual connections is indicated by arrows.}
%     \label{fig:GCN_res}
% \end{figure*}

Residual connection, also called skip-connection, was first
proposed in ResNet \cite{he2016deep} to solve the overfitting
issue of deep CNNs. Inspired by ResNet, recent
research has attempted to apply various residual connections to GCNs
to alleviate the over-smoothing problem, including CLN \cite{pham2017column} , Highway GCN \cite{rahimi2018semi} and JKNet \cite{xu2018representation} . However, the effectiveness of these methods is 
not satisfactory in large-scale graphs. In spite of this, several research still shows that residual connections are essential for deep GCNs, which can not only make GCNs have a more stable gradient, but also
partly alleviate the over-smoothing problem \cite{li2020deepergcn,chen2020simple}.

\begin{figure*}[htbp]
    \centering
    \includegraphics[width=\textwidth]{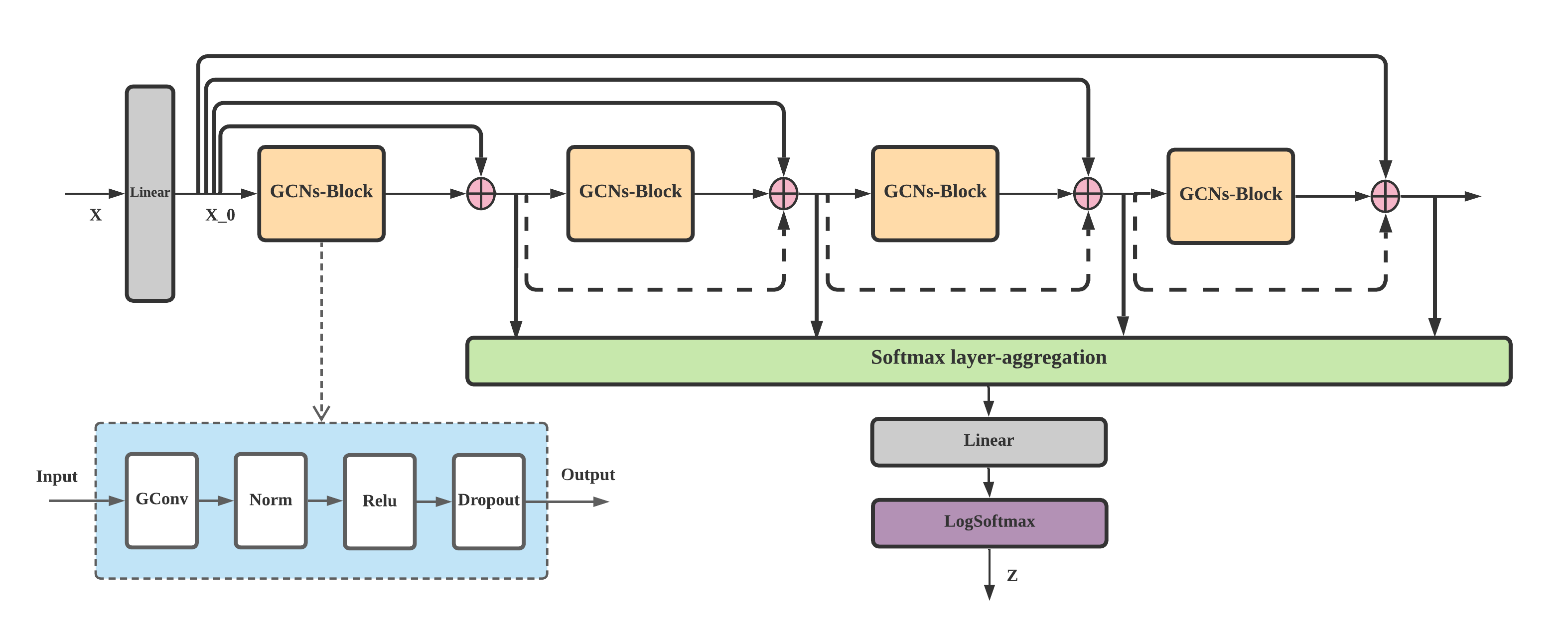}
    \caption{Overview of \textsl{GCN\_res Framework} with a 4-layer toy example. The GCNs-Block consists of four parts: \texttt{GCNsConv} layer, \texttt{Norm} layer, activation function, and \texttt{Dropout} unit. Data stream of residual connections is indicated by arrows.}
    \label{fig:GCN_res}
\end{figure*}

\begin{algorithm}[h]
\caption{GCN\_res Framework (i.e., forward propagation)}
\label{algo:framework}
\begin{algorithmic}[1]
\REQUIRE Graph $\mathcal{G}$; input feature matrix $\bm{X}$; depth K; weight matrices $\mathbf{W^{(k)}},\forall k \in \{1,...,K\}$; scale coefficients $\alpha,\beta$
\ENSURE Node representation $\bm{Z}$
\STATE $\bm{X^{(0)}}$ $\leftarrow$ Linear($\bm{X}$);
\FOR{k = 1, . . . ,K}
    \STATE $\bm{X^{(k)}}$ $\leftarrow$ GCNsConv$\left(\bm{X^{(k)}}, \mathcal{G}\right)$;
    \STATE $\bm{X^{(k)}}$ $\leftarrow$ Norm$\left(\bm{X^{(k)}}\right)$;
    \STATE $\bm{X^{(k)}}$ $\leftarrow$ Relu$\left(\bm{X^{(k)}}\right)$;
    \STATE $\bm{X^{(k)}}$ $\leftarrow$ Dropout$\left(\bm{X^{(k)}}\right)$;
    \STATE $\bm{X^{(k)}}$ $\leftarrow$ $\bm{X^{(k)}}$ + $\alpha$ * $\bm{X^{(0)}}$ + $\beta$ * $\bm{X^{(k-1)}}$;
\ENDFOR
\STATE $\mathbf{W^{(1)}}$,...,$\mathbf{W^{(K)}}$ $\leftarrow$ Softmax$\left(\mathbf{W^{(k)}},\forall k \in \{1,...,K\}\right)$;
\FOR{k = 1, . . . ,K}
    \STATE $\bm{X^{(k)}}$ $\leftarrow$ $\mathbf{W^{(k)}} \odot \bm{X^{(k)}}$
\ENDFOR
\STATE $\bm{Z}$ $\leftarrow$ Sum$\left(\bm{X^{(1)}},...,\bm{X^{(K)}}\right)$;
\STATE $\bm{Z}$ $\leftarrow$ Linear($\bm{Z}$);
\STATE $\bm{Z}$ $\leftarrow$ LogSoftmax($\bm{Z}$);
\end{algorithmic}
\end{algorithm}

In this paper, inspired by residual network, we
propose \textsl{GCN\_res Framework} by two main strategies in the forward propagation: (i) adaptive residual connections and initial residual connections; and (ii) softmax layer-aggregation. As an example, we show a 4-layer model which uses \textsl{GCN\_res Framework} in Fig. \ref{fig:GCN_res}. We also show the pseudo code of its forward propagation algorithm (Algorithm \ref{algo:framework}). In order to avoid the
expensive cost caused by high-dimension 
input feature matrix $\bm{X}$, we first reduce its dimension as $\bm{X^{(0)}}$ in the initialization phase by a \texttt{Linear} layer. Next, in each layer, \(\bm{X^{(k)}}\) passes through
\texttt{GCNsConv} layer, \texttt{Norm} layer, \texttt{Relu} function and \texttt{Dropout} unit
in turn. Then, we make $\bm{X^{(k)}}$ weighted summation with \(\bm{X^{(0)}}\) and \(\bm{X^{(k-1)}}\) according to the scale coefficients \(\alpha\) and \(\beta\) respectively. In our experiments, we fix
\(\alpha\) and \(\beta\) as hyper-parameters, which can also be obtained through learning. Finally, the node
representations \(\{\bm{X^{(1)}},...,\bm{X^{(K)}}\}\) of each layer are
aggregated by \textsl{softmax\ layer-aggregation} to obtain the final node representation $\bm{Z}$.

\subsubsection{Adaptive Residual Connections.} Inspired by DeeperGCN \cite{li2020deepergcn}, APPNP \cite{klicpera2018predict} and GCNII \cite{chen2020simple}, as shown in Algorithm \ref{algo:framework}, we apply adaptive residual connections and
initial residual connections to each layer in the forward
propagation to accelerate the convergence of GCNs and
alleviate the over-smoothing problem. By applying the above strategies to GCNs, the graph convolutional operator of our model can be rewritten as 

\begin{equation}
\label{equ:res}
\bm{X^{(k)}}=\sigma \left( Norm \left(GCNsConv\left(\bm{X^{(k-1)}},\mathcal{G}\right)\right)\right)+\alpha \cdot \bm{X^{(0)}} + \beta \cdot \bm{X^{(k-1)}},
\end{equation}

% \begin{equation}
% \label{equ:res}
% \bm{X^{(k)}}=\sigma \left( norm \left(\left(\widetilde D^{-\frac12}\widetilde A\widetilde D^{-\frac12} \bm{X^{(k-1)}} \mathbf{\Theta^{(k-1)}}\right)\right)\right)+\alpha \cdot \bm{X^{(0)}} + \beta \cdot \bm{X^{(k-1)}},
% \end{equation}

where $\sigma(\cdot)$ is an activation function, $Norm(\cdot)$ stands for the normalization functions, $GCNsConv(\cdot)$ stands for a convolutional layer, $\alpha$ and $\beta$ are two hyper-parameters.

\subsubsection{Softmax Layer-aggregation.} Inspired by CLN \cite{pham2017column} and JKNet \cite{xu2018representation}, we also propose softmax layer-aggregation. The difference between our method and jumping knowledge is that we use learnable weights which obey the softmax distribution in weighted summation for each layer instead of simply \textsl{sum} or \textsl{max}. The softmax layer-aggregation can be written as 

\begin{equation}
\label{equ:lg1}
\begin{aligned}
    softmax\left(\mathbf{W^{(1)}},...,\mathbf{W^{(K)}}\right), \\
    % \bm{Z} = \sum_{k=1}^{K} \mathbf{W^{(k)}} \odot \bm{X^{(k)}}
\end{aligned}
\end{equation}

\begin{equation}
\label{equ:lg2}
\begin{aligned}
    % softmax\left(\mathbf{W^{(1)}},...,\mathbf{W^{(K)}}\right) \\
    \bm{Z} = \sum_{k=1}^{K} \mathbf{W^{(k)}} \odot \bm{X^{(k)}},
\end{aligned}
\end{equation}

where $\mathbf{W^{(k)}}$ is a learnable weight matrix for weighted summation. The softmax layer-aggregation can effectively use the node representation
information of each layer to further alleviate the over-smoothing
problem.

In addition, Li $et\ al.$ \cite{li2020deepergcn} propose a pre-activated variant of residual connections for deep GCNs. In order to adjust GCN\_res Framework to the pre-activated version, we can put normalization and activation functions in front of \texttt{GCNsConv} layer. By integrating the above strategies together, GCN\_res Framework can be flexibly applied to
various baselines rather than limit to GCN, and it also provides a new possibility for the
future work of deep GCNs.
% In practice, we find that the model would achieve a good performance with $k=1,2$.

\subsection{Embedding Usage}
\bookmark[dest=\HyperLocalCurrentHref,level=2]{3.2 Embedding Usage}

% \begin{figure}[htbp]
%     \centering
%     \includegraphics[width=\textwidth]{Embedding.pdf}
%     \caption{\textsl{Embedding Usage} for GCNs. We merge input featrues with embedding to generate new features for GCNs.}
%     \label{fig:Embedding}
% \end{figure}

At present, models such as DeepWalk \cite{perozzi2014deepwalk} and Node2vec \cite{grover2016node2vec} can generate embeddings
for nodes in graphs through random walk. It has been confirmed that these embeddings contain rich graph structural information, however, the final accuracy of using these embeddings alone for node classification is generally low. This motivates the more general questions: \textsl{Are these embeddings really useful? Can GCNs use these embeddings?}

\begin{figure}[htbp]
    \centering
    \includegraphics[width=\textwidth]{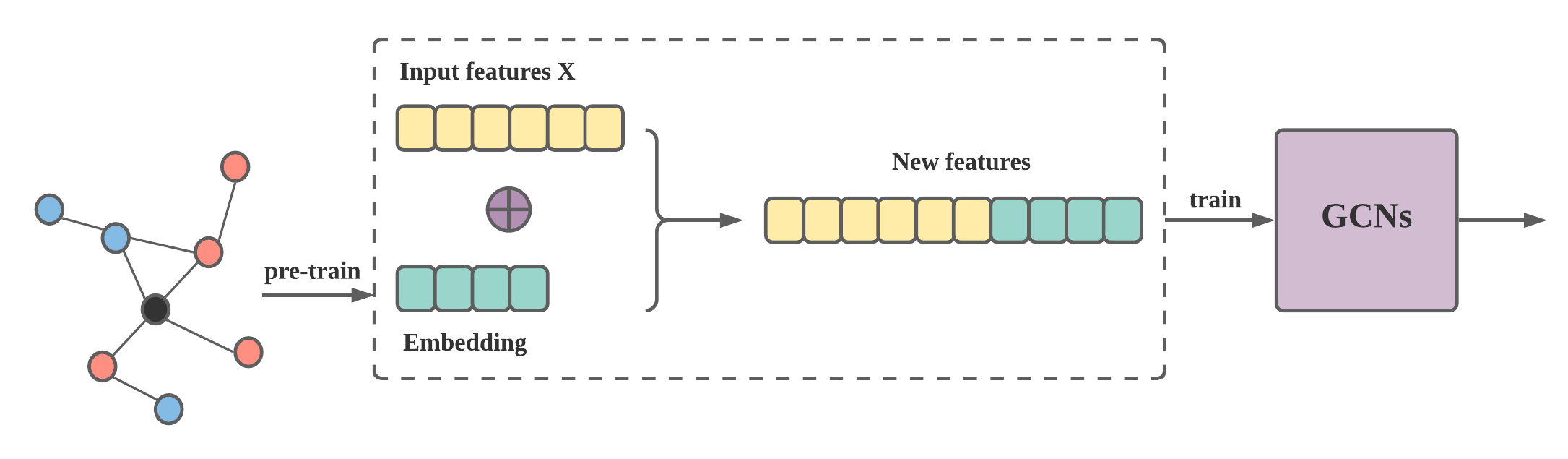}
    \caption{\textsl{Embedding Usage} for GCNs. We merge input features with embedding to generate new features for GCNs.}
    \label{fig:Embedding}
\end{figure}

In this work, we take an initial step towards answering the
questions above by proposing \textsl{Embedding Usage} to enhance node features. As shown in Fig. \ref{fig:Embedding}, we merge
the input feature matrix \(\bm{X}\) with embedding obtained by
pre-training, and then take the new feature matrix as input for GCNs. The merge process can be written as
\begin{equation} \label{equ:embed}
    \bm{X} = \mathbf{Merge}\left(\bm{X},\  Embeddings\right)  
\end{equation}

% \begin{equation}
% X = Merge\left(X, Metapath2vec\_Embedding\right)
% \end{equation}
where $Embeddings$ in Eq. (\ref{equ:embed}) can be Node2vec embedding in homogeneous graphs or MetaPath2vec \cite{dong2017metapath2vec} embedding in heterogeneous graphs. The $\mathbf{Merge}$ function can be $sum$, $concat$ (in our
experiments) or other functions. 

Theoretically, we can use any model to pre-train the embeddings. For
example, in heterogeneous graphs, we can also use TransE \cite{bordes2013translating} or DistMult \cite{yang2014embedding} 
embedding. Considering the cost and other factors, the model that uses to generate embeddings is
generally simpler than GCNs. These embeddings contain a lot of graph structural
information either in homogeneous or 
heterogeneous graphs, which can greatly improve the performance of GCNs.
Particularly, embeddings can play a pivotal role in the training process for heterogeneous graphs, where GCNs are severely under-fitting.

Besides, we can also use domain knowledge to
handcraft more features. For example, in molecular graph datasets, we can use chemical information and knowledge to handcraft more features for molecular representation learning, which is an interdisciplinary research direction.

%%% experiments
\section{Experiments}\label{experiments}
\bookmark[dest=\HyperLocalCurrentHref,level=1]{4 Experiments}
\subsection{Dataset}
\bookmark[dest=\HyperLocalCurrentHref,level=2]{4.1 Dataset}

\begin{table}[htbp] 
\caption{Statistics of the four OGB datasets for node classification tasks. The last column report the selected evaluation metrics in each dataset.}
\label{table:dataset-statistics}
\begin{center} 
% \begin{scriptsize}\begin{sc}
\renewcommand\tabcolsep{7.0pt}

\begin{tabular}{c|r|r|c|c}
% \toprule
\hline
\textbf{Dataset}  & \textbf{Nodes}  & \textbf{Edges} & \textbf{Classes} & \textbf{Metric} \\
% \midrule
\hline
\hline
ogbn-arxiv             & 169,343     & 1,166,243   &   40     & Accuracy \\
ogbn-mag         & 1,939,743 & 21,111,007 & 349 & Accuracy \\
ogbn-products          & 2,449,029   & 61,859,140	    & 47 &Accuracy \\
ogbn-proteins          & 132,534	    & 39,561,252    &  2  & ROC-AUC \\
% \bottomrule
\hline
\end{tabular} 
% \end{sc} \end{scriptsize} 

\end{center}
\end{table}

We conduct our experiments to evaluate the performance gain from two proposed tricks on four node classification datasets of OGB \cite{hu2020open}, including
\texttt{ogbn-arxiv}, \texttt{ogbn-mag}, \texttt{ogbn-products} and
\texttt{ogbn-proteins}. The statistics of each dataset is
shown in Table \ref{table:dataset-statistics}. Moreover, \texttt{ogbn-arxiv}, \texttt{ogbn-mag} and \texttt{ogbn-products} all contain node features,
and their evaluation metrics are Accuracy;
\texttt{ogbn-proteins} only contains edge features, and its
evaluation metric is ROC-AUC.

\subsection{Results and
Discussions}
\bookmark[dest=\HyperLocalCurrentHref,level=2]{4.2 Results and
Discussions}

In our experiments, we evaluate these tricks through ablation studies. In order to ensure the
fairness of the experiments, the hyper-parameters in the same dataset
are kept consistent. For GCN\_res, we set $\alpha$=0.2, $\beta$=0.7 in Eq. (\ref{equ:res}). In order to follow the rules of OGB leaderboard, 
the average and unbiased standard deviation of test accuracy are
taken over 10 different random seeds. Except for special
instructions, the best model for each dataset in our experiments can be
found in OGB leaderboard. The bold in the tables represents that 
the model is improved by our proposed tricks.

\subsubsection{A. OGBn-Arxiv.}
% ~\\
\texttt{ogbn-arxiv} is a homogeneous graph dataset. The results are presented in Table \ref{table:ogbn-arxiv-baseline}. We use GCN\_res to denote a 8-layer GCN with \textsl{GCN\_res
Framework}, which increases the test accuracy by 0.88\% compared to GCN. We also combine \textsl{GCN\_res Framework} with FLAG or C\&S
respectively to achieve further improvements. It is worth
mentioning that GCN\_res + C\&S\_v2 has a
1.39\% increase of the test accuracy compared to GCN. It achieves the best performance of models whose kernel are GCN and its variants, which confirms the
flexibility and robustness of \textsl{GCN\_res Framework}. Besides,
GCN\_res + C\&S\_v3, which uses the validation label in C\&S, is not announced in OGB leaderboard, however, its performance exceeds many GAT-based models. Compared with some baselines in OGB leaderboard, such as DeeperGCN \cite{li2020deepergcn}, GCNII \cite{chen2020simple} and UniMP \cite{shi2020masked}, our model achieves the best performance, which  confirms the effectiveness of \textsl{GCN\_res Framework} on \texttt{ogbn-arxiv}.

% \begin{table}[htbp] 
% \caption{The results of node classification on \texttt{ogbn-arxiv} in terms of Accuracy. The number in parentheses next to
% the model name indicates the number of layers.}
% \label{table:ogbn-arxiv}
% \begin{center}
% \renewcommand\tabcolsep{7.0pt}
% % \resizebox{0.9\linewidth}{!}{
% % \begin{scriptsize}\begin{sc}
% \begin{tabular}{l|c|c}
% % \toprule
% \hline
% \textbf{Model}  & \textbf{Test(\%)} & \textbf{Valid(\%)}  \\
% % \midrule
% \hline
% \hline
% GCN (3) \cite{kipf2016semi} & 71.74\std{0.29} & 73.00\std{0.17} \\
% GCN + FLAG (3) & 72.04\std{0.20} & 73.30\std{0.10} \\
% % \midrule
% \hline
% \textbf{GCN\_res} (8) & 72.62\std{0.37} & 73.69\std{0.21} \\
% \textbf{GCN\_res + FLAG} (8) & 72.76\std{0.24} & 73.89\std{0.12} \\
% % \textbf{GCN\_res + C\&S} (8) & 72.97\std{0.22} & 74.23\std{0.14} \\
% \textbf{GCN\_res} \textbf{+ C\&S\_v2} (8) & 73.13\std{0.17} &
% \textbf{74.45\std{0.11}} \\
% \textbf{GCN\_res} \textbf{+ C\&S\_v3} (8) & \textbf{73.91\std{0.14}} &
% 73.61\std{0.21} \\
% % \bottomrule
% \hline
% \end{tabular}
% % \end{sc} \end{scriptsize}
% % }
% \end{center} 
% \end{table}

\begin{table}[htbp] 
\caption{The results of node classification on \texttt{ogbn-arxiv} in terms of Accuracy. The number in parentheses next to
the model name indicates the number of layers.}
\label{table:ogbn-arxiv-baseline}
\begin{center}
\renewcommand\tabcolsep{7.0pt}
% \resizebox{0.9\linewidth}{!}{
% \begin{scriptsize}\begin{sc}
\begin{tabular}{l|c|c}
% \toprule
\hline
\textbf{Model}  & \textbf{Test(\%)} & \textbf{Valid(\%)}  \\
% \midrule
\hline
\hline
% GraphSAGE \cite{hamilton2017inductive} & 71.49\std{0.27} & 72.77\std{0.16} \\
MLP  & 55.50\std{0.23} & 57.65\std{0.12} \\
% GCN \cite{kipf2016semi} & 71.74\std{0.29} & 73.00\std{0.17} \\
GCN (3) \cite{kipf2016semi} & 71.74\std{0.29} & 73.00\std{0.17} \\
GCN + FLAG (3) & 72.04\std{0.20} & 73.30\std{0.10} \\
SIGN \cite{rossi2020sign} & 71.95\std{0.11} & 73.23\std{0.06} \\
DeeperGCN \cite{li2020deepergcn} & 71.92\std{0.16} & 72.62\std{0.14} \\
DAGNN \cite{liu2020towards} & 72.09\std{0.25} & 72.90\std{0.11} \\
JKNet \cite{xu2018representation} & 72.19\std{0.21} & 73.35\std{0.07} \\
GCNII \cite{chen2020simple}& 72.74\std{0.16} & --- \\
UniMP \cite{shi2020masked}& 73.11\std{0.20} & \textbf{74.50\std{0.05}} \\
% \midrule
\hline
\hline
\textbf{GCN\_res} (8) & 72.62\std{0.37} & 73.69\std{0.21} \\
\textbf{GCN\_res} + \textbf{FLAG} (8) & 72.76\std{0.24} & 73.89\std{0.12} \\
\textbf{GCN\_res} + \textbf{C\&S\_v2} (8) & 73.13\std{0.17} & 74.45\std{0.11} \\
\textbf{GCN\_res} + \textbf{C\&S\_v3} (8) & \textbf{73.91\std{0.14}} &
73.61\std{0.21} \\
% \bottomrule
\hline
\end{tabular}
% \end{sc} \end{scriptsize}
% }
\end{center} 
\end{table}

\begin{figure}[htbp]
\centering
\subfigure[MLP.]{
\includegraphics[width=4.5cm]{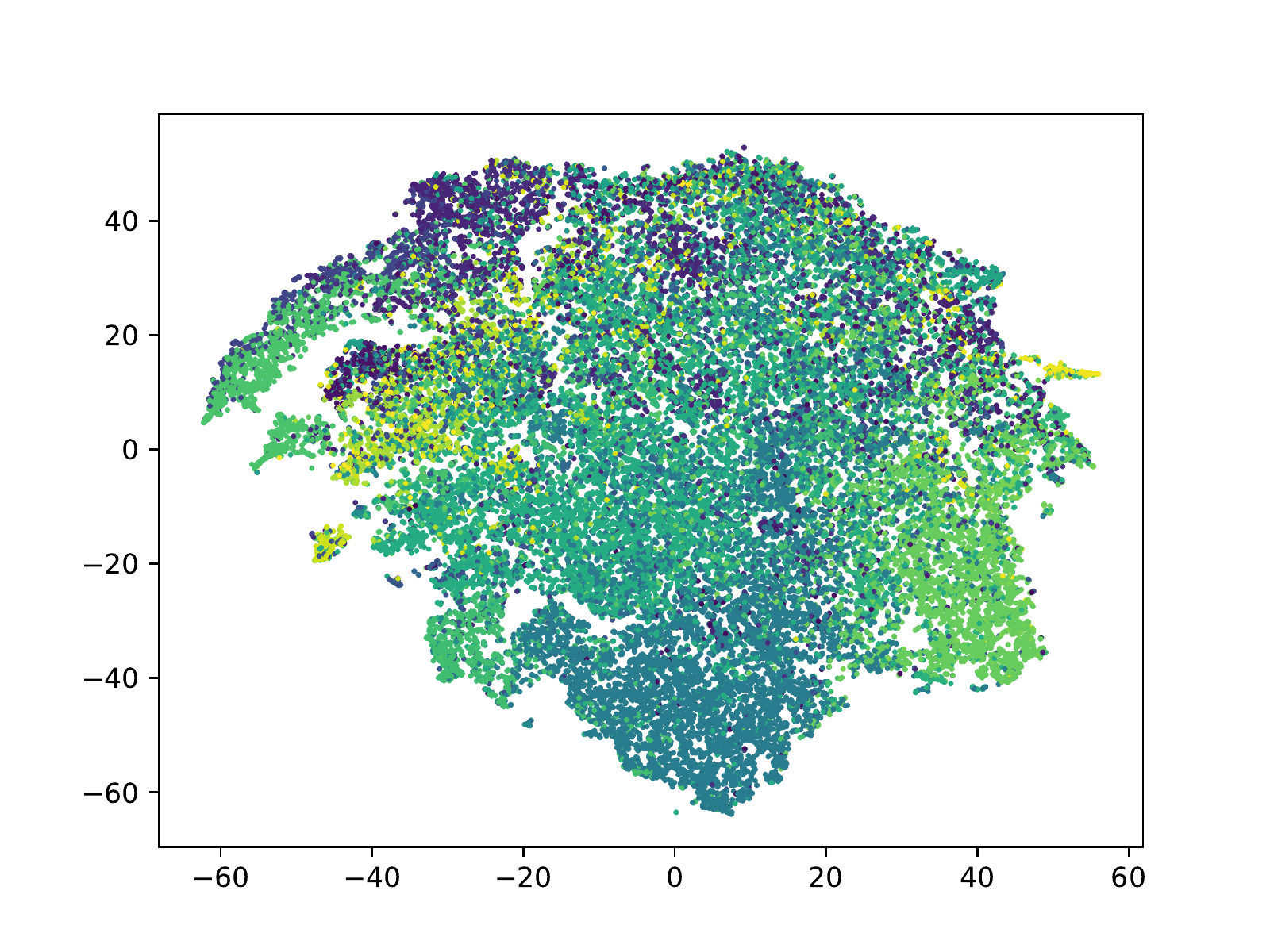}
%\caption{fig1}
}
\quad
\subfigure[GCN.]{
\includegraphics[width=4.5cm]{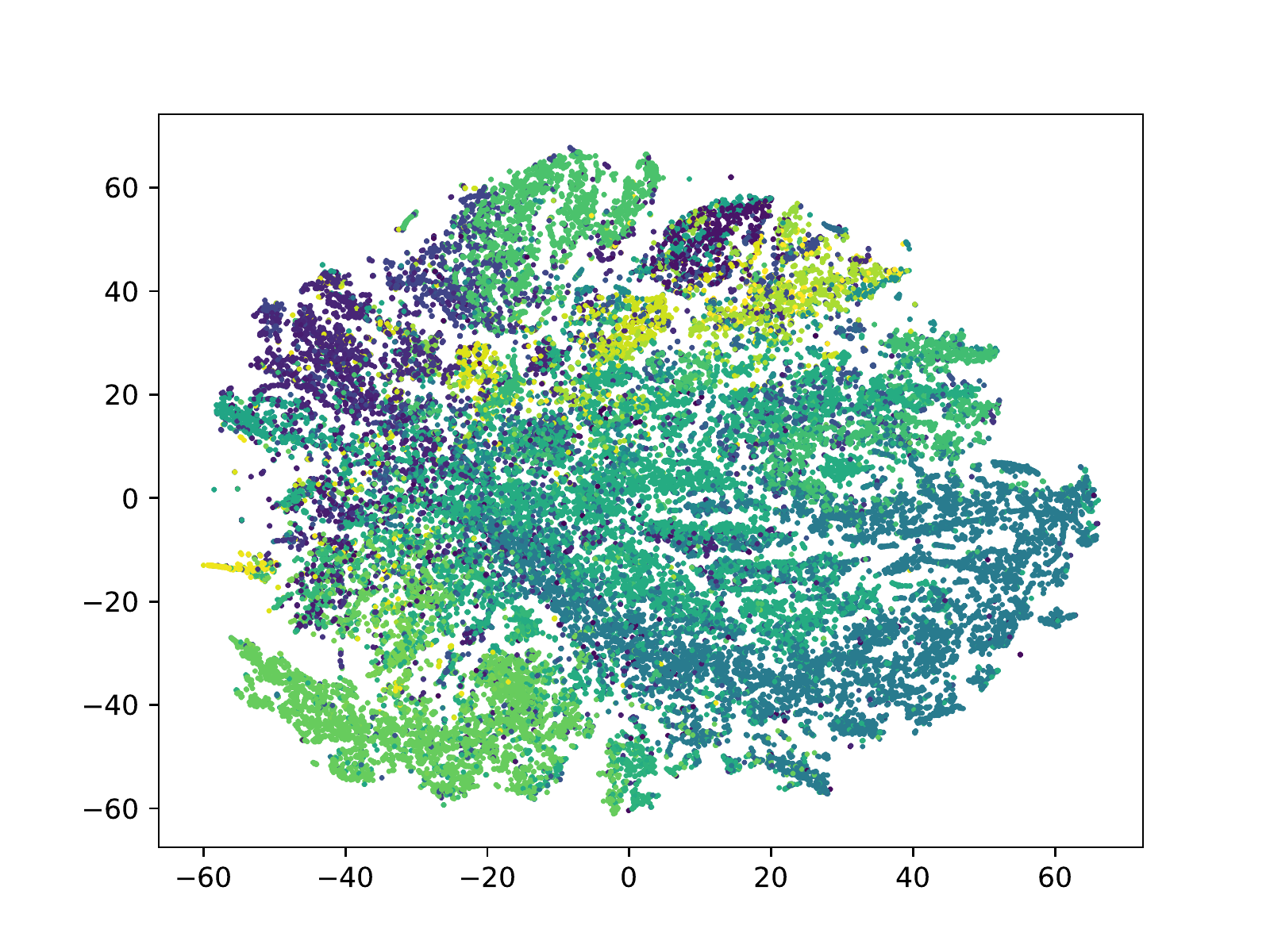}
}
\quad
\subfigure[GCN\_res+C\&S\_v2.]{
\includegraphics[width=4.5cm]{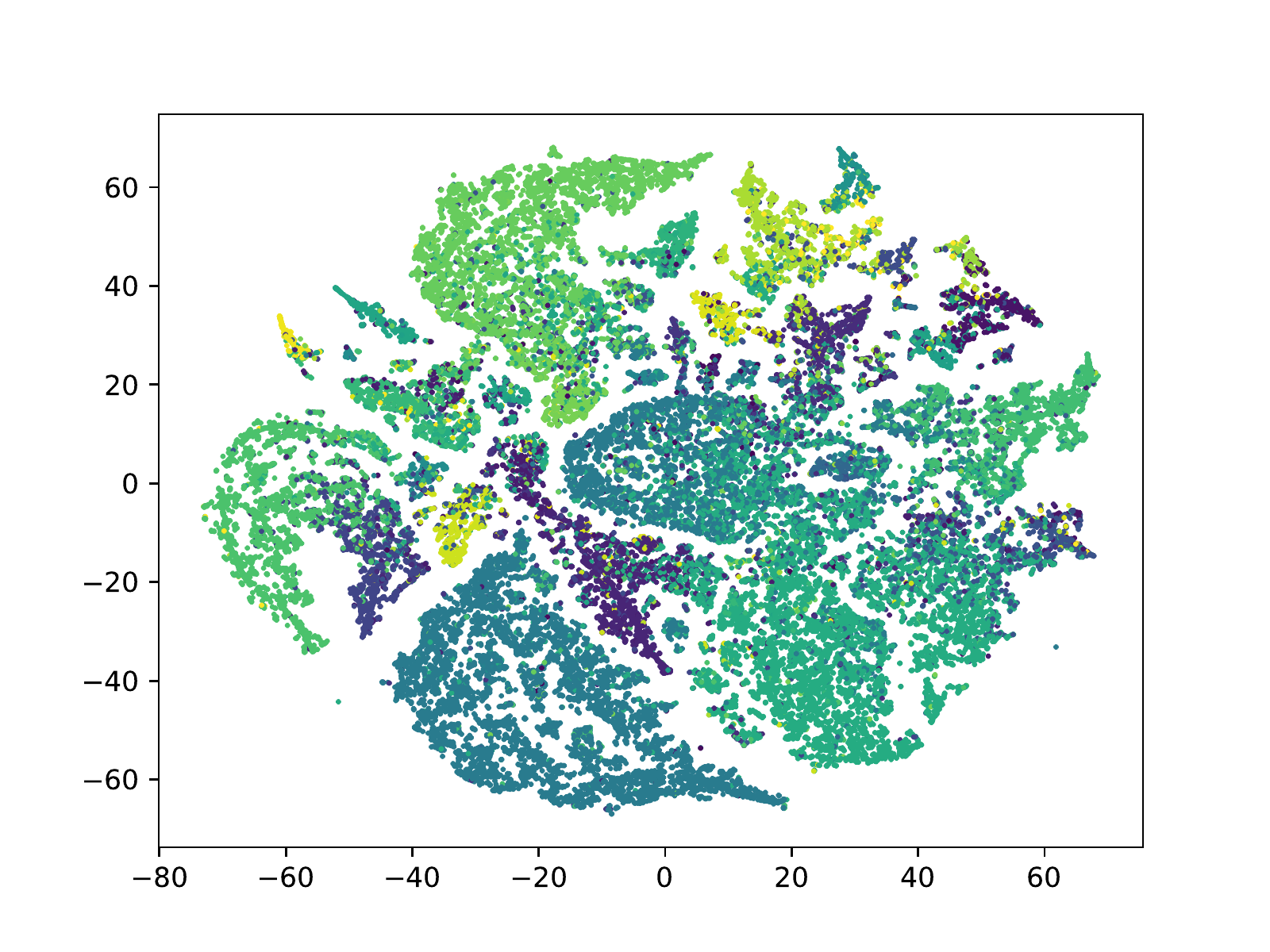}
}
\quad
\subfigure[GCN\_res+C\&S\_v3.]{
\includegraphics[width=4.5cm]{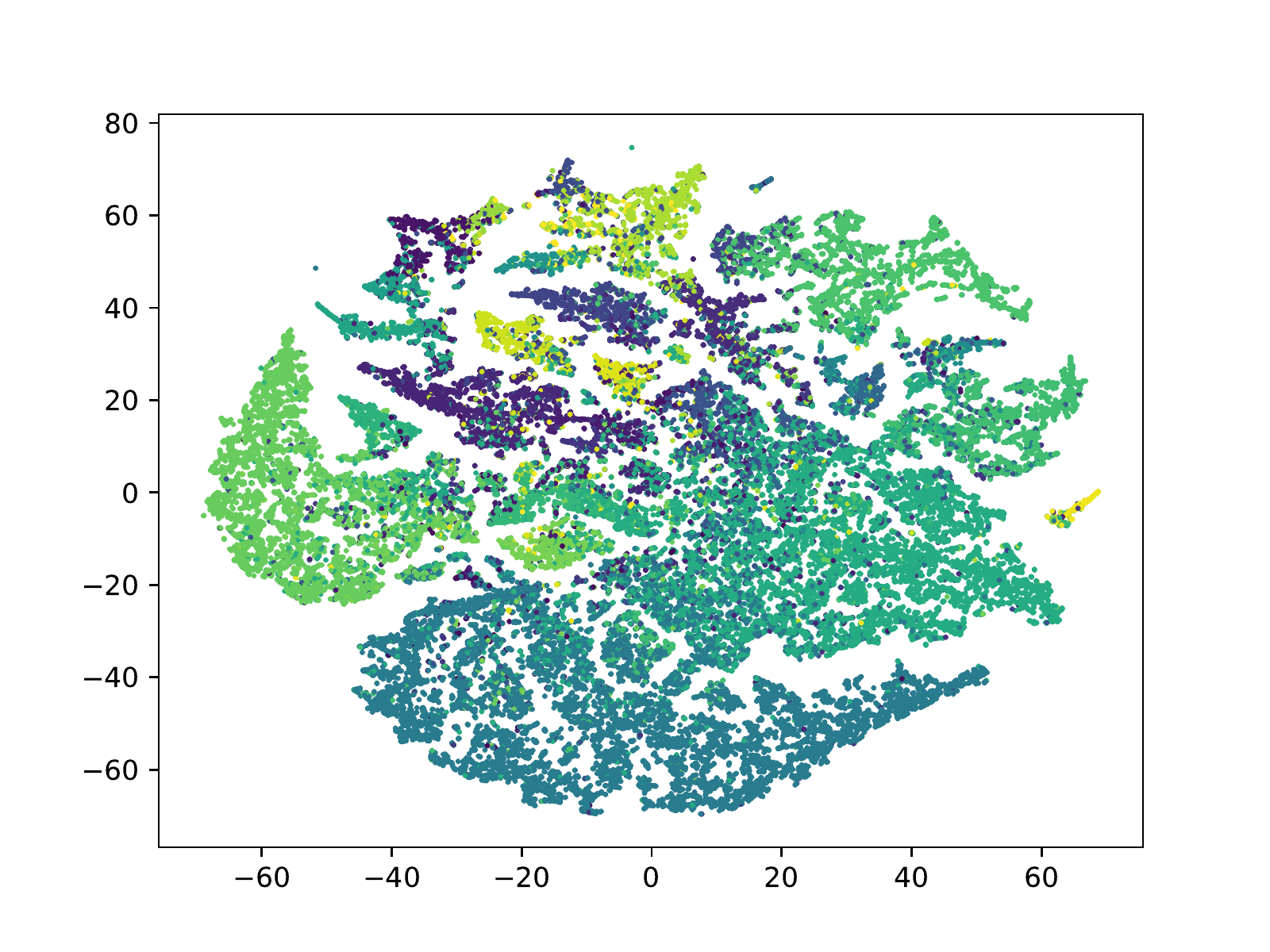}
}
\caption{t-SNE visualization results on \texttt{ogbn-arxiv}.}
\label{fig:tSNE}
\end{figure}

For a more intuitive comparison, we visualize the final test node representations by t-SNE. The results are shown in Fig. \ref{fig:tSNE}. It is obvious that our model can achieve more separated clusters than MLP and GCN, which demonstrates that \textsl{GCN\_res Framework} helps to learn more meaningful node embeddings.

\subsubsection{B. OGBn-Mag.}
% ~\\
\texttt{ogbn-mag} is a heterogeneous graph dataset. As shown
in Table \ref{table:ogbn-mag}, only by adding MetaPath2vec embedding for input features, the test accuracy of GraphSAINT increases 
by 2.15\%. On this basis, GraphSAINT + metapath2vec + FLAG even achieves the best performance of models whose kernel are R-GCN and its variants in OGB leaderboard. Interestingly, C\&S, which is an effective
method in homogeneous graphs, will make GCNs suffer an accuracy drop in heterogeneous graphs. The reason lies in the theoretical motivation for label propagation algorithm (LPA \cite{zhu2005semi})
--- adjacent nodes should have similar labels. However, there are
different types of nodes in heterogeneous graphs, and even
adjacent nodes will have completely different labels. Moreover, although FLAG can
improve the performance of R-GCNs, the effect is not obvious.

\begin{table}[htbp] 
\caption{The results of node classification on \texttt{ogbn-mag} in terms of Accuracy. }
\label{table:ogbn-mag}
\begin{center}
\renewcommand\tabcolsep{7.0pt}
% \resizebox{1.0\linewidth}{!}{
% \begin{scriptsize}\begin{sc}
\begin{tabular}{l|c|c}
% \toprule
\hline
\textbf{Model}  & \textbf{Test(\%)} & \textbf{Valid(\%)}  \\
% \midrule
\hline
\hline
GraphSAINT (R-GCN aggr) \cite{zeng2019graphsaint} & 47.51\std{0.22} & 48.37\std{0.26} \\
% \midrule
\hline
\hline
\textbf{GraphSAINT} + \textbf{metapath2vec} & 49.66\std{0.22} & 50.66\std{0.17} \\
\textbf{GraphSAINT} + \textbf{metapath2vec} + \textbf{C\&S} & 48.43\std{0.24} & 49.36\std{0.24} \\
\textbf{GraphSAINT} + \textbf{metapath2vec} + \textbf{FLAG} & \textbf{49.69\std{0.22}} &
\textbf{50.88\std{0.18}} \\
% \bottomrule
\hline
\end{tabular}
% \end{sc} \end{scriptsize}
% }
\end{center} 
\end{table}

Due to the complexity of heterogeneous graphs, the test and validation accuracy of GCNs on \texttt{ogbn-mag} are
relatively low. Many tricks used in homogeneous graphs cannot be
directly transferred to heterogeneous graphs. Nevertheless, it is clear that
embedding information is crucial for heterogeneous graphs, and how
to use this information rationally may become a future research direction.

% \begin{table}[htbp] 
% \caption{The results of node classification on \texttt{ogbn-mag} in terms of Accuracy. }
% \label{table:ogbn-mag}
% \begin{center}
% \renewcommand\tabcolsep{7.0pt}
% % \resizebox{1.0\linewidth}{!}{
% % \begin{scriptsize}\begin{sc}
% \begin{tabular}{l|c|c}
% % \toprule
% \hline
% \textbf{Model}  & \textbf{Test(\%)} & \textbf{Valid(\%)}  \\
% % \midrule
% \hline
% \hline
% GraphSAINT (R-GCN aggr) \cite{zeng2019graphsaint} & 47.51\std{0.22} & 48.37\std{0.26} \\
% % \midrule
% \hline
% \textbf{GraphSAINT + metapath2vec} & 49.66\std{0.22} & 50.66\std{0.17} \\
% \textbf{GraphSAINT + metapath2vec + C\&S} & 48.43\std{0.24} & 49.36\std{0.24} \\
% \textbf{GraphSAINT + metapath2vec + FLAG} & \textbf{49.69\std{0.22}} &
% \textbf{50.88\std{0.18}} \\
% % \bottomrule
% \hline
% \end{tabular}
% % \end{sc} \end{scriptsize}
% % }
% \end{center} 
% \end{table}

% attn_dropout=0.1, dropout=0.25, input_dropout=0.1, log_every=5, lr=0.01, n_epochs=1200, n_heads=6, n_hidden=80, n_layers=6

\subsubsection{C. OGBn-Products.}
% ~\\
\texttt{ogbn-products} is a homogeneous graph dataset. As shown in
Table \ref{table:ogbn-products}, using mini-batch algorithm based on neighbor sampling (NS) can
improve GraphSAGE's test accuracy slightly. The results also prove that sampling can
not only make training easier, but also be regarded as regularization
to improve the performance of GCNs. 
% The combination of BN and C\&S allows GraphSAGE to train with a more stable gradient while improving its performance greatly over time. 
Through the combination of BN, C\&S and Node2vec embedding, we raise GraphSAGE's test accuracy from 78.50\% to 81.54\%, which achieves the best performance of models whose kernel are GraphSAGE and its variants in OGB leaderboard. It is worth mentioning that Node2vec
embedding significantly improves the accuracy of GCNs, thus confirming the effectiveness and necessity of embedding
information once again.

\begin{table}[htbp] 
\caption{The results of node classification on \texttt{ogbn-products} in terms of Accuracy. }
\label{table:ogbn-products}
\begin{center}
\renewcommand\tabcolsep{5.0pt}
% \resizebox{\textwidth}{!}{
% \begin{scriptsize}\begin{sc}
\begin{tabular}{l|c|c}
% \toprule
\hline
\textbf{Model}  & \textbf{Test(\%)} & \textbf{Valid(\%)}  \\
% \midrule
\hline
\hline
Full-batch GraphSAGE \cite{hamilton2017inductive}& 78.50\std{0.14} & 92.24\std{0.07} \\
GraphSAGE w/NS & 78.70\std{0.36} & 91.70\std{0.09} \\
GraphSAGE w/NS + FLAG & 79.36\std{0.57} & 92.05\std{0.07} \\
% \midrule
\hline
\hline
\textbf{GraphSAGE w/NS} + \textbf{BN} + \textbf{C\&S} & 80.41\std{0.22} & 92.38\std{0.07} \\
\textbf{GraphSAGE w/NS} + \textbf{BN} + \textbf{C\&S} + \textbf{node2vec} & \textbf{81.54\std{0.50}} &
\textbf{92.38\std{0.06}} \\
% \bottomrule
\hline
\end{tabular}
% \end{sc} \end{scriptsize}
% }
\end{center} 
\end{table}

\subsubsection{D. OGBn-Proteins.}
% ~\\
\texttt{ogbn-proteins} is a homogeneous dataset. It
should be noted that \texttt{ogbn-proteins} does not have node features at first, so we use the summation edge features as initial node features in our experiments.
As shown in Table \ref{table:ogbn-proteins}, only using FLAG, the
test accuracy of GEN \cite{li2020deepergcn,li2019deepgcns} will decrease instead. By combining FLAG and Node2vec embedding, the test accuracy of
GEN increases by 1.21\%, which achieves the best performance of models whose kernel are GEN and its variants in OGB leaderboard. The Node2vec embedding enhances the node features,
thus improving the accuracy of GCNs.

\begin{table}[htbp] 
\caption{The results of node classification on \texttt{ogbn-proteins} in terms of ROC-AUC. }
\label{table:ogbn-proteins}
\begin{center}
\renewcommand\tabcolsep{7.0pt}
% \resizebox{0.9\linewidth}{!}{
% \begin{scriptsize}\begin{sc}
\begin{tabular}{l|c|c}
% \toprule
\hline
\textbf{Model}  & \textbf{Test(\%)} & \textbf{Valid(\%)}  \\
% \midrule
\hline
\hline
\textbf{GEN} \cite{li2020deepergcn,li2019deepgcns} & 81.30\std{0.65} & 85.74\std{0.53} \\
\textbf{GEN} + \textbf{FLAG} & 81.29\std{0.67} & 85.87\std{0.54} \\
\textbf{GEN} + \textbf{FLAG} + \textbf{node2vec} & \textbf{82.51\std{0.43}} & \textbf{86.56\std{0.37}} \\
% \bottomrule
\hline
\end{tabular}
% \end{sc} \end{scriptsize}
% }
\end{center} 
\end{table}

\subsubsection{E. Discussions.}
% ~\\
Since the inner relationship between tricks is not clear, there are few
papers that combine tricks with different motivations in their
experiments. In our experiments, we use two novel proposed tricks in
combination with other existing tricks, and discuss the
effects of various combination strategies in different datasets.
Kong \textsl{et al.} \cite{kong2020flag} point out that whether or not tricks are effective
depends on data distribution. Different graph structures will
greatly affect the expressive abilities of tricks.
Therefore, different tricks and their combinations should be used for
different datasets.

Through experiments on these datasets, we find that
\textsl{GCN\_res Framework} in the citation network \texttt{ogbn-arxiv}
successfully alleviates over-smoothing issue of GCNs, which can also 
easily integrate various types of tricks due to its good generality and
flexibility. The Node2vec or MetaPath2vec embedding can
make GCNs have an obvious improvement in most scenarios so that we strongly
recommend you to use embeddings in GCNs. Besides, normalization and dropout are
generally used to stabilize the gradient of GCNs and
prevent the problem of overfitting. In addition, C\&S
has a positive effect on homogeneous graphs, but is not applicable in
heterogeneous graphs. FLAG can slightly improve the performance of
GCNs on most datasets, but this is not a large increase.

The combination of various tricks can achieve better performance gain
than a single trick, for example, GCN\_res + C\&S\_v3, GraphSAINT + metapath2vec + FLAG, GraphSAGE w/NS + BN + C\&S + node2vec and GEN + FLAG + node2vec all
achieve the best performance in our experiments. Therefore, exploring the inner relationship
between different types of tricks and making rational use of them may be
a future research direction.

%%% conclusion
\section{Conclusion}\label{conclusion}
\bookmark[dest=\HyperLocalCurrentHref,level=1]{5 Conclusion}

In this paper, we first summarize the mini-batch training process and existing effective tricks of GCNs for node classification tasks from a practical perspective. In addition, we also propose
two novel tricks: \textsl{GCN\_res Framework} and \textsl{Embedding Usage}, which can improve various GCNs via residual network and pre-trained embedding. Experiments
on OGB datasets show that the combination of these tricks can
greatly improve the performance of GCNs. Due to the low cost and
flexibility, tricks will play an important role in practical
applications of GCNs in the future.

\newpage

\bibliographystyle{splncs04}
\bibliography{mybibliography}
%





% LNCS Homepage, \url{http://www.springer.com/lncs}. Last accessed 4
% \end{thebibliography}

\end{document}